# Integrated Scheduling Model for Arrivals and Departures in Metroplex Terminal Area

Tonghe Li,[*] Jixin Liu,[†] Hao Jiang,[‡] Weili Zeng[§] and Lei Yang[**]
*Nanjing University of Aeronautics and Astronautics, 211106 Nanjing, People's Republic of China*

In light of the rapid expansion of civil aviation, addressing the delays and congestion phenomena in the vicinity of metroplex caused by the imbalance between air traffic flow and capacity is crucial. This paper first proposes a bi-level optimization model for the collaborative flight sequencing of arrival and departure flights in the metroplex with multiple airports, considering both the runway systems and TMA (Terminal Control Area) entry/exit fixes. Besides, the model is adaptive to various traffic scenarios. The genetic algorithm is employed to solve the proposed model. The Shanghai TMA, located in China, is used as a case study, and it includes two airports, Shanghai Hongqiao International Airport and Shanghai Pudong International Airport. The results demonstrate that the model can reduce arrival delay by 51.52%, departure delay by 18.05%, and the runway occupation time of departure flights by 23.83%. Furthermore, the model utilized in this study significantly enhances flight scheduling efficiency, providing a more efficient solution than the traditional FCFS (First Come, First Served) approach. Additionally, the algorithm employed offers further improvements over the NSGA II algorithm.

[*] Postgraduate Student, College of Civil Aviation; also State Key Laboratory of Air Traffic Management System and Technology, 210007 Nanjing, People's Republic of China.
[†] Associate Professor, College of Civil Aviation; also State Key Laboratory of Air Traffic Management System and Technology, 210007 Nanjing, People's Republic of China; larryljx66@nuaa.edu.cn.
[‡] Ph.D. Student, College of Civil Aviation; also State Key Laboratory of Air Traffic Management System and Technology, 210007 Nanjing, People's Republic of China; jianghao_cca@nuaa.edu.cn.
[§] Associate Professor, College of Civil Aviation; also State Key Laboratory of Air Traffic Management System and Technology, 210007 Nanjing, People's Republic of China; zwlnuaa@nuaa.edu.cn.
[**] Associate Professor, College of Civil Aviation; also State Key Laboratory of Air Traffic Management System and Technology, 210007 Nanjing, People's Republic of China.

## Nomenclature

$AF_j$ = the arrival fix flight $j$ passes

$ARP_i$ = $i$ th airport in the multi-airport system

$C_{j,k}$ = the clearance separation for departure flights $j$ and $k$ passing identical departure fix when they are ready for taking off

$CR$ = the time expended by the landed flight in crossing the departure runway adjacent to its landing runway

$D$ = the time taken by the departure flight from take-off acceleration to clearing from the runway

$DF_j$ = the departure fix flight $j$ passes

$F^{arr}$ = the set of arrival flights in the multi-airport system

$F^{dep}$ = the set of departure flights in the multi-airport system

$F^{arr}_{ARP_i}$ = the set of arrival flights in $ARP_i$

$F^{dep}_{ARP_i}$ = the set of departure flights in $ARP_i$

$H^{arr}$ = ATC handover separation at arrival fixes

$H^{dep}$ = ATC handover separation at departure fixes

$h^{arr}_j$ = handover altitude of arrival flight $j$

$h^{dep}_j$ = handover altitude of departure flight $j$

$mps_j$ = the maximum position shift for flight $j$

$N$ = number of airports in the multi-airport system

$O_j$ = the time offset when the order of flight $j$ in the queue of landing flights is advanced or delayed by one place

$n_{j_{LD}}$ = scheduled landing order of arrival flight $j$

$n^{opt}_{j_{LD}}$ = optimized landing order of arrival flight $j$

$P$ = binary variable indicates the peak state of the multi-airport system

| | | |
|---|---|---|
| $RW_j$ | = | the runway flight $j$ lands or takes off |
| $t_{j_{ARR}}$ | = | planned arrival time of arrival flight $j$ |
| $t_{j_{ARR}}^{opt}$ | = | optimized arrival time of arrival flight $j$ |
| $t_{j_{DEP}}$ | = | planned departure time of departure flight $j$ |
| $t_{j_{DEP}}^{opt}$ | = | optimized departure time of departure flight $j$ |
| $t_{j_{LD}}$ | = | planned landing time of arrival flight $j$ |
| $t_{j_{LD}}^{opt}$ | = | optimized landing time of arrival flight $j$ |
| $t_{j_{TO}}$ | = | planned take-off time of departure flight $j$ |
| $t_{j_{TO}}^{opt}$ | = | optimized take-off time of departure flight $j$ |
| $t_{j_{APP}}^{smu}$ | = | the simulated approach time of arrival flight $j$ |
| $t_{j_{CL}}^{smu}$ | = | the simulated climbing time of departure flight $j$ |
| $VA_j$ | = | time expended by landing flight $j$ in vacating from the landing runway |
| $W_{j,k}^{LD}$ | = | wake separation between landing flight $j$ and landing flight $k$ |
| $W_{j,k}^{TO}$ | = | wake separation between take-off flight $j$ and take-off flight $k$ |

## I. Introduction

THE rapid expansion of the civil aviation industry has resulted in significant delays and congestion at the metroplex terminal area, mainly due to an imbalance in air traffic patterns. In the context of the metroplex terminal area, two distinct operational coupling issues have been identified. The first pertains to the coupling between arrival and departure flights, while the second concerns the coupling between flights originating from different airports. The limited airspace and runway resources in the metroplex terminal area are shared by different flights, resulting in operational inefficiency and increased flight delays and congestion. In response to this challenge, the FAA NextGen[1], ICAO ASBU [2], and EUROCONTROL SESAR [3] have proposed the use of flight sequencing and scheduling methods to rationally allocate time and space resources in a metroplex terminal area, intending to improve operational efficiency in the TMA.

Compared to TMAs with only one airport, the air traffic capacity and flow of a multi-airport system (MAS) appear to be significantly enhanced, thereby increasing the complexity of the flight scheduling problem. For example, in a metroplex terminal area comprising two airports, A and B, several scenarios may arise during periods of high traffic. Firstly, flights from A, either climbing or approaching, may crowd the airspace resources of B, creating a potential conflict. Secondly, changes in landing times at A may result in further tightening of runway slots due to the conflict with B. In light of the inter-airport operational coupling at both the temporal and spatial levels, metroplex terminal area-oriented sequencing and scheduling is regarded as a solution to the uneven allocation of spatial and temporal resources within the MAS. This is achieved by adjusting the relative order of flights and exchanging flight time slots within each airport to avoid potential conflicts in spatial and temporal resources. As a result, delays and congestion in the metroplex terminal area can be alleviated. In a metroplex terminal area, the potential for sequence adjustment exists concerning the airport runways and the waypoints within the airspace. In this case, air traffic controllers can direct flights to pass through most waypoints at flexible altitudes. They can guide them near the waypoints to avoid potential conflicts. However, the handover fixes, including the arrival and departure fixes, play the role of the entrances and exits of TMA, and every flight must pass the handover fix at the end of the climbing segment and the beginning of the approaching segment. Additionally, only one or two fixed handover altitudes are at the aforementioned handover fixes. Similar to airport runways, the handover fixes load most delays and congestion in the metroplex terminal area. Wang [4] and Ma [5] took into account the ATC handover separation constraints at the handover fixes in their researches about flight sequencing problems in the metroplex terminal areas.

While a framework for flight collaborative sequencing between runways and handover fixes has been established, notable areas require further optimization. Paramount among these is prioritizing arrival flights, especially in a metroplex terminal area environment. In the actual operation of air traffic control, to coordinate the conflict between arrival and departure flights, the arrival flights often need to hold in the air or follow radar vectoring to a specific arrival path. Such an operation has the additional consequence of prolonging flight time and increasing fuel consumption while also burdening air traffic controllers. Optimizing approach flights is particularly important when conducting collaborative sequencing of arrival and departure flights. This optimization directly reduces the waiting time for passengers on board. It also facilitates the release of airspace resources. Additionally, it decreases fuel consumption. Ultimately, this improves the operational efficiency of the entire TMA. Although Jiang et al. [6,7] considered prioritizing arrival flights in a single-airport-single-runway scenario, the application of this concept remains in the exploratory stage in a metroplex terminal area environment. This is particularly pertinent given the necessity for prioritizing arrival flights in such an environment due to the coupling of arrival and departure flights at different airports.

Following this, the extant flight sequencing model does not fully consider the relative order of the front and rear aircraft at the same airport and handover fix during climb or approach. This deficiency may result in conflicts and delays due to the necessity for sequence adjustments. Lastly, most extant studies, derived from actual ATC operations, posit that flights at each airport must adhere to a fixed handover altitude when traversing the handover fix. This assumption constrains the effective utilization of airspace resources, and the fixed altitude allocation increases the complexity of air traffic control and potentially results in additional delays.

To address the issues above, the model proposed in this paper employs a bi-level programming model based on the approach priority concept within a metroplex terminal area. The upper and lower models are applied to optimize the approach and departure flights, respectively. Secondly, the model proposed in this paper assumes that all front and rear aircraft at the same airport and handover fix are not permitted to exchange their relative order with each other. This avoids additional delays and conflicts caused by unnecessary order exchanges that may occur during the sequencing process. Finally, this article allocates handover altitudes through the staggered assignment of handover altitudes based on the chronological order of flights passing through the same handover fix during the data processing phase rather than assigning fixed handover altitudes based on airports. Specifically, for a handover fix that offers one or two handover altitudes, the method ensures that consecutive flights passing through the same fix are assigned different altitudes (handover altitude 1 and handover altitude 2) whenever possible. This differentiation of handover altitudes between front and rear aircraft minimizes unnecessary horizontal handover separation, thereby enhancing the efficiency and reliability of flight scheduling within the Metroplex Terminal Area.

The main contributions of the paper are as follows:

1) A bi-level programming model is utilized in the metroplex terminal area, where the upper level optimizes the scheduling of arrival flights, while the lower level focuses on optimizing departure flights. This design ensures that the overall sequencing model has higher optimization priority for the arrival flights and, at the same time, mitigates the impact of the operational coupling of the arrival and departure flights on the optimization model.

2) We introduce an innovative flight sequencing model that maintains the relative order of departure flights departing from the same airport and arriving at the same handover fix in the metroplex terminal area. By structuring the sequencing process in this manner, our model significantly streamlines the workflow for air traffic controllers, enhancing operational simplicity while upholding the highest flight safety standards. This method minimizes the need for frequent and often disruptive sequence adjustments, thereby preventing potential conflicts and delays. Consequently, the overall efficiency of air traffic management is preserved, if not improved, by reducing the instances of unnecessary interventions. Our approach leverages the inherent patterns in flight trajectories to optimize traffic flow, offering a robust solution for complex airspace environments.

3) The paper introduces a novel method for allocating handover altitudes through the staggered assignment of handover altitudes. Unlike traditional approaches that assign fixed altitudes based on airports, the proposed method allocates altitudes based on the sequence of flights passing through the same handover fix. Specifically, when a handover fix offers two possible handover altitudes, the method alternates the assignment between consecutive flights, ensuring that adjacent flights are assigned different altitudes (handover altitude 1 and handover altitude 2) whenever possible. This staggered assignment minimizes unnecessary handover separation, thereby preventing potential delays that may arise when air traffic control (ATC) enforces additional separation due to fixed handover altitudes.

The following is a description of the organization of the chapters in this paper: Section 2 presents a review of the literature on the current state of flight scheduling research; Section 3 introduces the collaborative sequencing model for metroplex bi-level programming of departure flights proposed in this paper; Section 4 presents the algorithms used in the model; Section 5 presents the results of experiments conducted on the model and algorithms, and Section 6 concludes with a summary and outlook.

## II. Literature review

The research on flight sequencing is divided into two categories: single-airport-oriented flight sequencing research and multi-airport-oriented flight sequencing research. In the single-airport flight sequencing problem, Dear [8] proposed the Constrained Position Shift (CPS) concept in 1976. This reduced the solution space of the model and improved the solution time and feasibility of the model. Subsequently, Hu et al. [9] introduced the concept of

Rolling Horizon Control (RHC), which notably reduced the model solution size and convergence difficulty while maintaining comparable optimization performance to previous solution methods. This advancement successfully integrated the flight scheduling and sequencing problem into the pre-tactical phase of air traffic flow management. Later, Hu and Di Paolo [10] devised a 0-1 binary genetic algorithm tailored to address the arrival flight sequencing and scheduling challenges, building upon the RHC concept. Thereafter, Hu and Di Paolo [11] further refined the binary genetic algorithm, customized for the arrival flight sequencing and scheduling problem, by optimizing the design of the underlying operators to maintain chromosome diversity while representing chromosome matrices using binary variables. The CPS constraints and the RHC concept are the two significant contributions that originate from the study of single-airport flight sequencing and are referenced by most of the metroplex terminal area flight sequencing studies.

The current studies on the metroplex terminal area are divided into three categories: arrival flight sequencing [4,5,12–16], departure flight sequencing [17–22], and arrival-departure flight collaborative sequencing [23–30]. In the studies of arrival flight sequencing in a metroplex terminal area, Ma et al. [5] employed runway-handover-fix collaborative sequencing for approach flights in a metroplex terminal area. They set the objective as airport fairness and economy in a multi-airport system and solved it using the NSGA II algorithm. Salehipour et al. [12] established a mixed integer goal programming model describing the arrival flights sequencing problem. They developed a hybrid metaheuristic applying a simulated annealing framework to solve the problem efficiently. Cao et al. [13] developed a hot-start greedy algorithm for the inbound flight scheduling problem, improving the solution's performance. Zhang et al. [14] proposed a criteria selection method for determining the optimization objective of the arrival flight sequencing problem. Jiang et al. [15] further considered the intersections of approach flights from different airports within a multi-airport system and sequenced them. Murça et al. [16] introduced the concept of alternative approach routes and generated outputs that could be converted into valid flight instruction execution recommendations. Wang et al. [4] structured paired approach sequencing model at closely spaced parallel runways based on point merge scenario.

In the studies of departure flight sequencing in a metroplex terminal area, Wang et al. [17] proposed the use of a shared segment as well as shared waypoint constraints for flights departing from different airports in a multi-airport system. They then sequenced these flights and solved the problem using a tabu search algorithm. Montoya et al. [18] used a dynamic programming algorithm to optimize flight delays and runway capacity bi-objectively. Liu et al. [19]

abstracted the metroplex terminal area departure flight sequencing problem into a two-stage no-wait traveler problem. Zhong et al. [20] addressed the issue of departure flight sequencing in a multi-airport system by formulating it as a two-phase, no-waiting-traveler problem. They considered the scenario in which arrival flights from Airport A may restrict the operation of departure flights from Airport B during peak approach times. To address this, they designed a tabu search algorithm to find an optimal solution. Sandamali et al. [21] considered demand uncertainty brought by departure and en-route speed uncertainty in their departure flight sequencing problems model. Ma et al. [22] fully considered slot restrictions and holding queue capacity in their model and developed a simulated annealing algorithm to solve the problem.

In the studies of arrival-departure flight collaborative sequencing in a metroplex terminal area, Chandrasekar et al. [23] proposed a model for assigning and sequencing runways to arrival and departure flights without constraining runway configurations. They solved the model using the branch-bounding approach. In their model, Sölveling et al. [24] considered uncertain aircraft availability on the runway. Shi et al. [25] constructed a collaborative flight sequencing problem and solved it using a simulated annealing algorithm. Ahmed et al. [26] established a collaborative optimization model of flight sequencing and runway configuration for a multi-airport system. They proposed a collaborative and co-evolutionary genetic algorithm to solve the problem.

Notwithstanding the advances in previous research, several under-explored issues remain concerning optimizing flight scheduling in metroplex terminal areas within multi-airport systems. Firstly, although studies have commenced to examine the collaborative effect between runways and handover fixes, there is a paucity of in-depth exploration of the relationship between ATC handover separation and handover altitude at handover fixes and its impact on the operational efficiency of the entire system. Additionally, most current studies on flight sequencing in metroplex terminal areas prioritize delay reduction as the optimization objective while neglecting to consider the respective optimization demands of arrival and departure flights under different traffic scenarios. Moreover, most studies employ static or semi-static methodologies to address the flight scheduling problem. However, the dynamic characteristics of flight traffic over time in actual operating environments are frequently overlooked. This may result in the optimization scheme's lack of adaptability and robustness in practical applications. Accordingly, this study addresses the aforementioned research gaps by proposing a novel collaborative sequencing method for flights in metroplex terminal areas. This method is based on an in-depth analysis of the constraints associated with handover fixes and a dynamic optimization model that employs bi-level programming. The objective is to enhance the efficiency of flight scheduling while simultaneously ensuring the stability and reliability of the system.

### III. Methodology

**A. Problem Description**

In a multi-airport system (MAS), when a flight is situated within the TMA, the flight proceeds through either a departure climb or an approach descent. The departure flight commences at the runway and culminates at the handover fix (departure point), while the approach flight initiates at the handover fix and progresses towards the runway. Both arrival and departure flights must adhere to stringent altitude restrictions at the commencement and conclusion of their respective journeys. At the outset, the runway altitude is set to zero, while the altitude of the handover fix is established at one or two predetermined handover altitudes. It is not mandatory for an arriving or departing flight to traverse all the remaining waypoints in the Standard Terminal Arrival (STAR) or Standard Instrument Departure (SID) procedures. Instead, flights may pass through these waypoints at flexible altitudes following the circumstances. Therefore, this paper will focus on the collaborative sequencing between airport runways and handover fixes within the metroplex terminal area.

This paper is dedicated to optimizing departure flights' take-off times and arrival flights' crossing arrival fix times for each multi-runway airport at a metroplex terminal area. Subsequently, the arrival flight will perform the approach segment after crossing the approach fix and before landing. Similarly, the departure flight will perform the climbing segment after taking off and before crossing the departure fix. The arrival and departure flights are coupled to each other on the runway. Specifically, for an arrival flight on one runway and a departure flight on the adjacent parallel runway, which are closely spaced, there are departure-arrival flight wake spacing constraints and arrival-departure flight crossing runway constraints. While flights that may be at different airports but have the same arrival or departure fix are coupled together at the handover fixes, in particular, such flights have ATC handover separation constraints. Furthermore, the objectives and results of model optimization in a metroplex terminal area will vary depending on the specific traffic scenario. This is because the metroplex terminal area as a whole, as well as the relative capacity-traffic conditions of each airport, will be considered. This paper will discuss the traffic scenarios in a metroplex terminal area to address this issue. Fig. 1 shows a schematic representation of the problem addressed in this paper.

## B. Model Formulation

Arrival flights play a crucial role in reducing the workload of air traffic controllers. They also enhance the efficiency of airlines and increase airport throughput. Given this importance, this paper introduces bi-level programming for modeling and solving related issues [6,7,31]. Bi-level programming doubles the maximum solving capacity of the heuristic algorithm, with the bi-level model focusing on optimizing the same type of flights, thereby reducing overall complexity. The upper model optimizes arrival flights, while the lower model optimizes departure flights. In each iteration, the upper model first solves the solution that optimizes the arrival flights. Then, the solution is passed to the lower model, which adjusts the optimization of the departure flights. In contrast, the incoming solution of the upper model, i.e., the arrival flights, remains unchanged. The subsequent iteration commences, and the departure flights solved by the lower model are transferred to the upper model, which modifies the global solution to optimize the arrival flights. This process continues until the objective function values of both the upper and lower models reach convergence, at which point the solution is concluded.

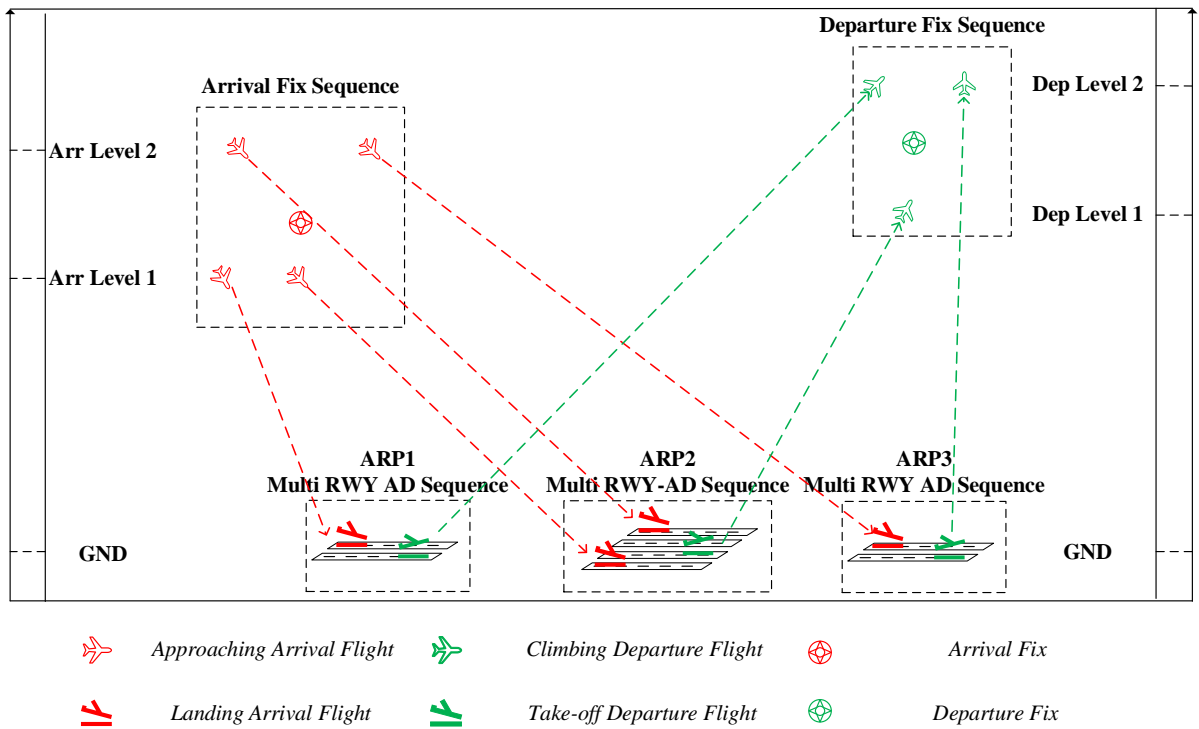

**Fig. 1 Schematic diagram of collaborative arrival-departure flight sequencing in a metroplex terminal area**

In the study of collaborative runway-handover-fix arrival and departure flight sequencing in the metroplex terminal area, the interaction between the two airports exists in the portion of the handover-fix sequencing. In contrast, the interaction between the arrival and departure flights exists on multiple runways at each airport. As mentioned earlier, the runway time (take-off/landing time) and the time to cross the handover fix (departure handover/arrival handover time) of the arrival flights are bridged by a time-of-flight simulation.

*1. Upper-Level Optimization Model for Arrival Flights*

Therefore, this paper sets two optimization objectives for the upper-level model, for the total sequence offsets of the arrival flights when the multi-airport system is in the peak hours and for the total delays of the arrival flights when the multi-airport system is in the non-peak hours. The optimization objective of the upper-level model is shown in Eq. (1):

$$\min R^U = \sum_{i=1}^{N} \sum_{j \in F_{ARP_i}^{arr}} P \times (n_{j_{LD}}^{opt} - n_{j_{LD}}) + (1-P) \times (t_{j_{ARR}}^{opt} - t_{j_{ARR}}) \quad (1)$$

where the binary variable $P$ is set to 1 when the multi-airport system is operating at its peak state and zero otherwise. The term "$ARP_i$" represents the $i$ th airport in the multi-airport system for the $j$ th flight, with a total of $N$ airports in the multi-airport system. The term "$F_{ARP_i}^{arr}$" denotes the set of all arrival flights in $ARP_i$. $t_{j_{ARR}}$ represents the scheduled arrival time of a flight and $t_{j_{ARR}}^{opt}$ represents the arrival time of an optimized flight. $n_{j_{LD}}$ and $n_{j_{LD}}^{opt}$ represent the landing order of arrival flights before and after optimization, respectively. It is evident that in the absence of an order change, the sub-objective function will remain constant at zero. Conversely, if an order change occurs, the sub-objective function will become a multiple of two.

The model constraints are listed as follows:

1) Wake Separation for Consecutive Arrival Flights:

According to the ATC separation requirement [32], the landing time between consecutive arrival flights must maintain the wake separation for arrival-arrival flights $W_{j,k}^{LD}$, if they use the same runway. Specifically, for consecutive arrival flights $j$ and $k$, if the same runway is used, the wake separation needs to be maintained between the rear aircraft and the front aircraft, as shown in Eq. (2):

$$t_{k_{LD}}^{opt} > t_{j_{LD}}^{opt} + W_{j,k}^{LD}, \forall j,k \in F_{ARP_i}^{arr}, RW_j = RW_k, t_{k_{LD}} > t_{j_{LD}} \quad (2)$$

where $t_{j_{LD}}$ represents the planned landing time for flight $j$, $RW_j$ represents the runway flight $j$ lands, $W_{j,k}^{LD}$ represents the wake separation between flight $j$ and $k$. If flight $j$ and $k$ are both arrival flights at the same airport and use the same runway $RW_j = RW_k$, and $j$ is scheduled to land earlier than $k$, then $k$ needs to maintain a wake separation with $j$.

As in Eq. (2), each constraint has a specific range, indicated after the constraint equation. There are three types of constraint ranges. Firstly, the set to which the two flights belong is denoted as $F$. Secondly, the interrelationships between the two flights concerning the use of the runway, designated as $RW$, and the handover fix, which can be either an arrival fix (notated as $AF$) or a departure fix (notated as $DF$), form the second type of constraint. Moreover, the third type involves the interrelationships between the two flights' scheduled times for taking off, landing, or passing through a handover fix. These three types of relationships qualify for a particular constraint that is in effect only for specific flights.

2) Runway Separation for Departure-Arrival Flights:

Based on the Air Traffic Control (ATC) separation requirements, when an arrival flight $k$ and a departure flight $j$ are using a pair of closely spaced parallel runways for landing and taking off, respectively, specific timing rules must be followed. If the target landing time of flight $k$ is later than the target take-off time of flight $j$, then flight $k$ cannot be cleared to land until flight $j$ has safely departed and cleared of the runway, as shown in Eq. (3):

$$t_{k_{LD}}^{opt} > t_{j_{TO}}^{opt} + D, \forall j \in F_{ARP_i}^{dep}, \forall k \in F_{ARP_i}^{arr}, \| RW_j, RW_k \|, t_{k_{LD}} > t_{j_{TO}} \tag{3}$$

where $F_{ARP_i}^{dep}$ represents the set of departure flights at airport $i$, $\| RW_j, RW_k \|$ represents flight $j$ and flight $k$ use a pair of parallel runways in close proximity, $D$ represents the time taken by the departure flight from take-off acceleration to overrun the end of the runway or to complete a turn in order to clear of the runway. $t_{j_{TO}}$ represents the planned take-off time for flight $j$, and $t_{j_{TO}}^{opt}$ represents the optimized take-off time for flight $j$.

3) Constant Relative Sequence for Same-Path Flights

It is assumed that two consecutive front and rear aircraft (later referred to as "same-path two aircraft") approaching from the same arrival fix and landing at the same airport have exchanged their order. This scenario does not affect flights coming from other arrival fixes. This does not contribute to the efficiency of the model as a whole. For the front aircraft of the two flights, this results in an unproductive delay and congestion of the slot resources of the more rearward flights approaching from the same runway. Therefore, this paper assumes that the two airplanes on the same path may not exchange their relative order. This constraint reduces unnecessary delays and improves efficiency. Fig. 2 illustrates the schematic diagram of this constraint.

This constraint can be formulated as follows:

$$t^{opt}_{k_{LD}} > t^{opt}_{j_{LD}}, \forall j,k \in F^{arr}_{ARP_i}, AF_j = AF_k, t_{k_{ARR}} > t_{j_{ARR}} \tag{4}$$

where $AF_j$ represents the arrival fix that flight j passes. This constraint represents that if both flight *j* and *k* belong to the set of arrival flights at the same airport $F^{arr}_{ARP_i}$, and they pass the same arrival fix ($AF_j = AF_k$), and the planned arrival time for j precedes that for k ($t_{k_{ARR}} > t_{j_{ARR}}$), then during the approach segment, it is imperative that the relative sequences of j and k remain unaltered. The constraint explicitly states that *j* is optimized to land earlier than k ($t^{opt}_{k_{LD}} > t^{opt}_{j_{LD}}$).

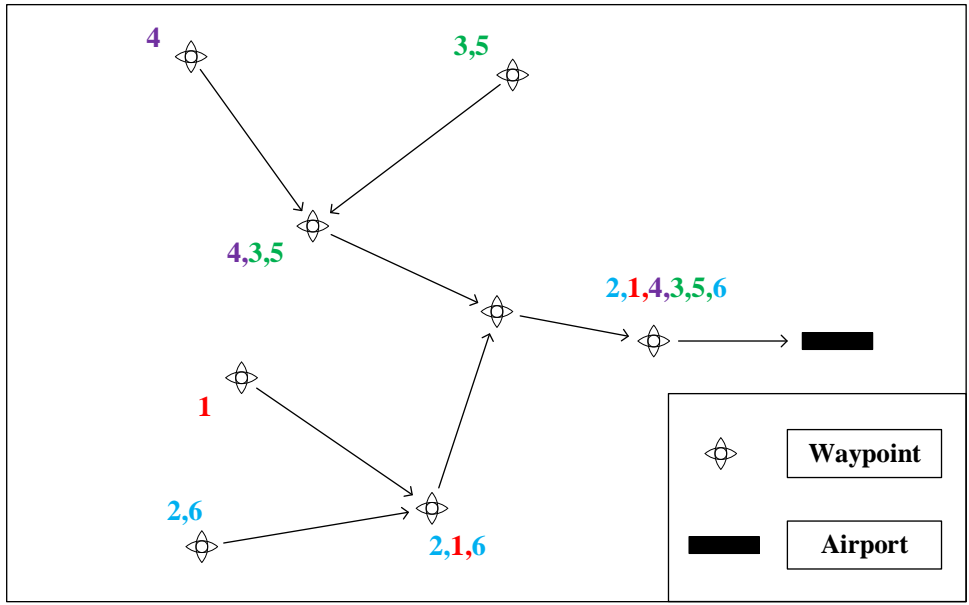

**Fig. 2 Constant Relative Sequence for Same-Path Flights**

4) ATC Handover Separation for Same-Handover-Altitude Arrival Flights

Traditional approaches assign each available approach handover altitude at an arrival fix to specific airports to alleviate controller workload within a multi-airport system. Accordingly, a constraint is established as follows: sufficient horizontal spacing must be maintained at the arrival fix when arrival flights are assigned the same handover altitude.

However, this methodology can inadvertently introduce unnecessary delays. To address this issue, during the data-processing phase, the model implements a staggered assignment of handover altitudes for flights passing through the same handover fix with multiple available handover altitudes. Specifically, when a handover fix offers two distinct handover altitudes, the model alternates the assignment between consecutive flights based on their chronological order. This ensures that adjacent flights are assigned different handover altitudes (handover altitude 1 and handover altitude 2) whenever possible. By differentiating the handover altitudes of consecutive flights, the model minimizes the need for additional horizontal separation mandated by air traffic control (ATC), thereby reducing potential delays and enhancing overall scheduling efficiency within the Metroplex Terminal Area. Fig. 3 illustrates the schematic diagram of this constraint.

This constraint can be formulated as follows:

$$t_{k_{ARR}}^{opt} > t_{j_{ARR}}^{opt} + H^{arr}, \forall j,k \in F^{arr}, AF_j = AF_k, h_j^{arr} = h_k^{arr} \tag{5}$$

where $h_j^{arr}$ represents the handover altitude of flight j. Suppose flight k is the first flight after flight j to pass through the same arrival fix. In that case, their handover altitudes are differentiated as much as possible during the data processing phase to avoid unnecessary horizontal handover separation. The term "$F^{arr}$" represents the set of arrival flights in the whole MAS, while "$H^{arr}$" represents ATC handover separation at arrival fixes. If flight j and flight k share the same arrival fix and handover altitude, they must maintain ATC handover separation.

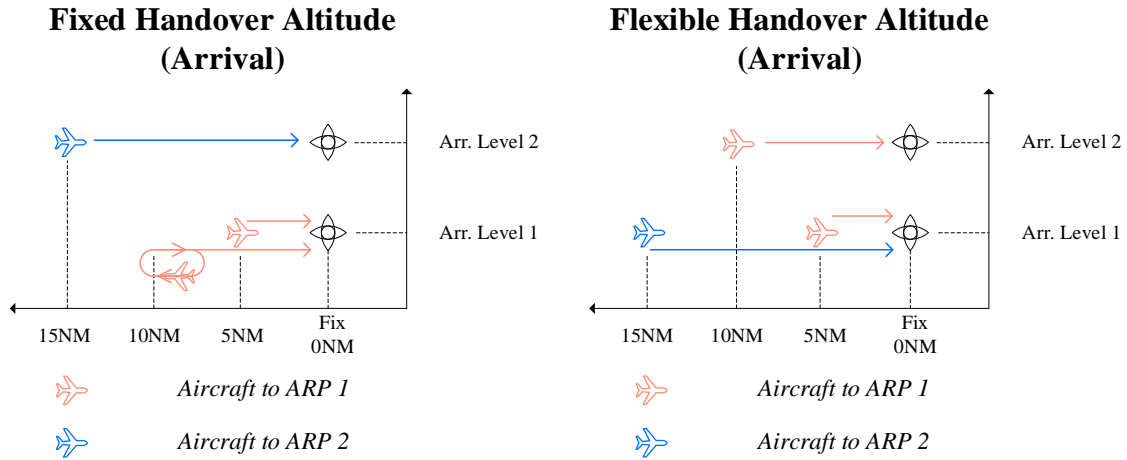

**Fig. 3 Flexible Handover Altitudes for Arrival Flights**

5) Runway Crossing Separation for Arrival-Departure Flights

According to the operation rules of proximity parallel runways, when flights are using a pair of parallel runways that are close to each other, with one runway for arrivals and the other for departures, the departure flight must wait for the arrival flight to cross the departure runway before the departure flight can take off if the following conditions are true. Firstly, the target take-off time of the departure flight is later than the target landing time of the arrival flight. Moreover, the airport is not equipped with end-around taxiways. Fig. 4 shows the schematic diagram of this constraint.

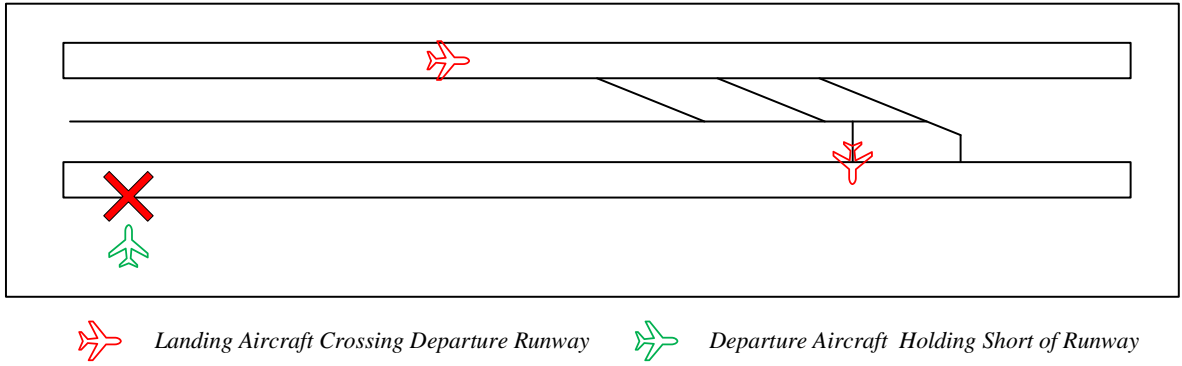

⇾ *Landing Aircraft Crossing Departure Runway*   ⇾ *Departure Aircraft Holding Short of Runway*

**Fig. 4 Runway Crossing Separation for Arrival-Departure Flights**

This constraint can be formulated as follows:

$$t^{opt}_{k_{TO}} > t^{opt}_{j_{LD}} + VA_j + CR, \forall j \in F^{dep}_{ARP_i}, \forall k \in F^{arr}_{ARP_i}, \| RW_j, RW_k \|, t_{k_{TO}} > (t_{j_{LD}} + VA_j) \tag{6}$$

where $CR$ denotes the time expended by the landed flight in crossing the departure runway adjacent to its landing runway. $VA_j$ denotes the time expended by landing flight $j$ in vacating from the landing runway. If the scheduled landing time of the arrival flight $j$ plus the time it takes for flight $j$ to vacate the runway is still earlier than the scheduled departure time of the departure flight $k$, then certain steps must be followed. First, the departure flight $k$ must wait for the arrival flight $j$ to land and vacate the runway. Then, k still needs to wait for $j$ to cross the take-off runway. Only after $j$ has crossed can $k$ take off.

6) Constrained Position Shift Constraint for Arrival Flights

For each arrival flight, adjusting a larger order leads to an enormous controller workload, so a constrained position shift (CPS) constraint is set [10]. This constraint prevents the model from making significant changes to the order of arrival flights, which would result in longer extra flight time and longer extra workload for the controllers. This constraint can be formulated as follows:

$$t_{j_{ARR}} - mps_j \times O_j \leq t^{opt}_{j_{ARR}} \leq t_{j_{ARR}} + mps_j \times O_j, \forall j \in F^{arr} \tag{7}$$

where $mps_j$ is a constant representing the maximum position shift for $j$, and in this research, this value has been taken as 2. $O_j$ is the time offset when the order of flight $j$ in the queue of landing flights is advanced or delayed by one place. This constraint implies that the landing time of landing flight $j$ can only be adjusted within the range $[-mps_j \times O_j, mps_j \times O_j]$.

7) Time-of-Flight Simulation Equation Constraint for Arrival Flights

Time-of-Flight Simulation links arrival time and landing time for arrival flights by outputting the flight time of certain segments. This paper uses the XGBoost [33] method to simulate the flight time of arrival flights and flight time of departure flights. All approach and departure flights are grouped based on the runways and handover fixes. And for each group of trajectories, the features are extracted according to the methods in the literature [34,35], and the flight times in the ADS-B data are used as labels for training.

This constraint can be formulated as follows:

$$t_{j_{LD}}^{opt} = t_{j_{ARR}}^{opt} + t_{j_{APP}}^{smu}, \forall j \in F^{arr} \tag{8}$$

where $t_{j_{APP}}^{smu}$ indicates the simulated approach time of flight $j$, linking optimized arrival time $t_{j_{ARR}}^{opt}$ and optimized landing time $t_{j_{LD}}^{opt}$.

*2. Lower-Level Optimization Model for Departure Flights*

The lower-level model is set up to solve the departure flight optimization problem. For the reasons mentioned in the introduction to the upper model, the optimization is performed for the Runway Occupation Time (ROT) of the departure flights when the multi-airport system is in the peak hours and for the Total Delay of the departure flights when the multi-airport system is in the non-peak hours. The optimization objective of the lower-level model is shown in Eq. (9):

$$\min R^L = P \times (\max_{j \in F^{dep}}(t_{j_{TO}}^{opt}) - \min_{j \in F^{dep}}(t_{j_{TO}}^{opt})) + (1-P) \times \sum_{i=1}^{N} \sum_{j \in F_{ARP_i}^{dep}} (t_{j_{TO}}^{opt} - t_{j_{TO}}) \tag{9}$$

where $F^{dep}$ represents the set of departure flight in MAS, $F_{ARP_i}^{dep}$ presents the set of departure flights at airport $i$, $t_{j_{TO}}$ represents the planned take-off time of flight $j$, and $t_{j_{TO}}^{opt}$ represents optimized take-off time of flight $j$.

Constraints of the lower level are listed as follows:

1) Wake Separation for Consecutive Departure Flights:

According to the ATC separation requirement, between consecutive departure flights, if the runways are the same, the wake separation needs to be followed:

$$t^{opt}_{k_{TO}} > t^{opt}_{j_{TO}} + W^{TO}_{j,k}, \forall j,k \in F^{dep}_{ARP_i}, RW_j = RW_k, t_{k_{TO}} > t_{j_{TO}} \tag{10}$$

where $W^{TO}_{j,k}$ indicates wake separation between departure flights $j$ and $k$.

2) Clearance Separation for Departure Flights

According to the ATC separation requirement, between departure flights, if both flights pass an identical departure fix, clearance separation needs to be obeyed:

$$t^{opt}_{k_{TO}} > t^{opt}_{j_{TO}} + C_{j,k}, \forall j,k \in F^{dep}_{ARP_i}, DF_j = DF_k, t_{k_{TO}} > t_{j_{TO}} \tag{11}$$

where $DF_j$ indicates departure fix that flight $j$ passes, and $C_{j,k}$ represents the clearance separation for departure flights passing identical departure fix when ready for taking off.

3) Runway Separation for Departure-Arrival Flights:

As with the upper model constraint (3), if an arrival flight and a departure flight use a set of close parallel runways and the target landing time of the arrival flight is later than the target departure time of the departure flight, then the arrival flight needs to wait for the departure flight to clear of runway before it can land:

$$t^{opt}_{k_{LD}} > t^{opt}_{j_{TO}} + D, \forall j \in F^{dep}_{ARP_i}, \forall k \in F^{arr}_{ARP_i}, \| RW_j, RW_k \|, t_{k_{LD}} > t_{j_{TO}} \tag{12}$$

4) Constant Relative Sequence for Same-Path Flights

In accordance with constraint (4) of the upper model, departure flights that consecutively leave from the same airport and are destined for the same handover fix must maintain their original order during the climbing phase:

$$t^{opt}_{k_{DEP}} > t^{opt}_{j_{DEP}}, \forall j,k \in F^{dep}_{ARP_i}, DF_j = DF_k, t_{k_{TO}} > t_{j_{TO}} \tag{13}$$

where $t^{opt}_{j_{DEP}}$ indicates the optimized departure time of flight $j$ when $j$ passes the departure fix, and $t_{j_{DEP}}$ indicates the schemed departure time of flight $j$.

5) ATC Handover Separation for Same-Handover-Altitude Departure Flights

In the same way as the upper model constraint (5), if the departure fix and departing handover altitude of departure flights j and k remain the same, they shall maintain ATC handover separation at departure fix, as Fig. 5.

The constraint can be formulated as follows:

$$t^{opt}_{k_{DEP}} > t^{opt}_{j_{DEP}} + H^{dep}, \forall j,k \in F^{dep}, DF_j = DF_k, h^{dep}_j = h^{dep}_k, t_{k_{DEP}} > t_{j_{DEP}} \tag{14}$$

where $H^{dep}$ represents ATC handover separation at departure fixes, and $h^{dep}_j$ represents handover altitude of flight $j$.

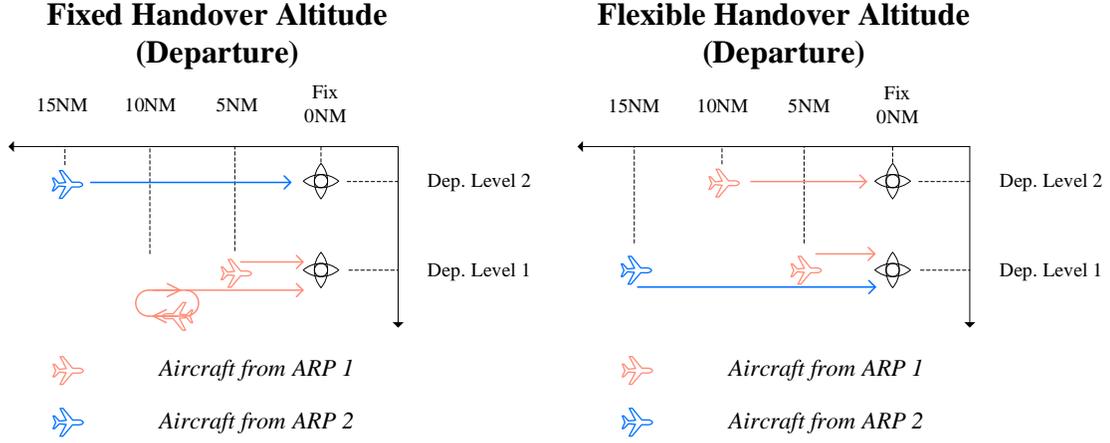

Fig. 5 Flexible Handover Altitude for Departure Flights

6) Runway Crossing Separation for Arrival-Departure Flights

Same as the upper model constraint (6), when flights are using a pair of parallel runways that are close to each other, with one runway for arrivals and the other for departures, the departure flight must wait for the arrival flight to cross the departure runway before the departure flight can take off if the following conditions are true. Firstly, the target take-off time of the departure flight is later than the target landing time of the arrival flight. Moreover, the airport is not equipped with end-around taxiways.

$$t^{opt}_{k_{TO}} > t^{opt}_{j_{LD}} + VA_j + CR, \forall j \in F^{dep}_{ARP_i}, \forall k \in F^{arr}_{ARP_i}, \| RW_j, RW_k \|, t_{k_{TO}} > (t_{j_{LD}} + VA_j) \tag{15}$$

7) Constrained Position Shift Constraint for Departure Flights

Unlike arrival flights, departure flights can only delay their take-off time instead of advancing.

$$t_{j_{TO}} \leq t^{opt}_{j_{TO}} \leq t_{j_{TO}} + mps_j \times O_j, \forall j \in F^{dep} \tag{16}$$

8) Time-of-Flight Simulation Equation Constraint for Departure Flights

Time-of-Flight Simulation links departure time and take-off time for departure flights by outputting flight time of specific segment:

$$t^{opt}_{j_{DEP}} = t^{opt}_{j_{TO}} + t^{smu}_{j_{CL}}, \forall j \in F^{dep} \tag{17}$$

where $t_{j_{CL}}^{smu}$ indicates the simulated climbing time of flight $j$, linking optimized departure time $t_{j_{DEP}}^{opt}$ and optimized take-off time $t_{j_{TO}}^{opt}$.

## IV. Algorithm Design

Heuristic evolutionary algorithms show significant advantages in solving NP-Hard problems such as flight scheduling problems. The proposed model determines the optimization objective through different MAS traffic scenarios. This design can reflect the difference between the peak and non-peak states of the multi-airport system and circumvents the problem of difficulty in converging the solution of the bi-level multi-objective evolutionary algorithm. In this paper, we design upper-level and lower-level models with their objectives. These objectives are defined for different scenarios. The models are then input into a bi-level single-objective heuristic evolutionary algorithm. Our study uses two main optimization algorithms: the bi-level Elitist Genetic Algorithm, known as bi-EGA, and the bi-level Strengthened Elitist Genetic Algorithm, referred to as bi-SEGA. Additionally, we employ a bi-level baseline algorithm for comparison. This baseline is the bi-level simple genetic algorithm known as bi-GA. To evaluate performance, we compare the bi-GA with bi-EGA and bi-SEGA. In this case, the overall architecture of the bi-XX algorithm is a structure of two identical instances of the XX algorithm, which pass the results of their respective solutions to each other and iterate consecutively [22]. The performance of these algorithms is compared through experimental methods. The goal is to identify the bi-level single-objective heuristic evolutionary algorithm that performs optimally.

During the execution of the algorithm, various interval-based constraints need to be handled. These constraints can be transformed in the following way. During each iteration of the upper and lower levels of the algorithm, the population decides on whether or not the constraint is violated after going through crossover and mutation. Individuals in the population store genes as a list of arriving times for an arrival flight or take-off times for a departure flight and merge the current individual with the incoming optimal individual of the other level; accordingly, the individual's arrival/departure schedules are extracted and merged with the optimal individual of the other level's departure/arrival schedules, which yields, at the time of the current individual's determination, the complete runway schedule of flights within the MAS along with the full schedules of all handover fixes. Then extract the runway information, landing and take-off airport information, and handover fix information for each flight. Based on the matching degree of the above three information between flights, the runway times of the flights belonging to the current individual in this table are judged with interval constraints, and the individuals violating the interval constraints are eliminated.

The proposed algorithm is shown in Fig. 6. The input data is ADS-B data of departure and arrival flights. The specific steps of the algorithm are listed as follows:

**Step 1** Extract the required data for the algorithm from ADS-B data, including the landing and take-off times of arrival and departure flights and the time of passing the handover fixes.

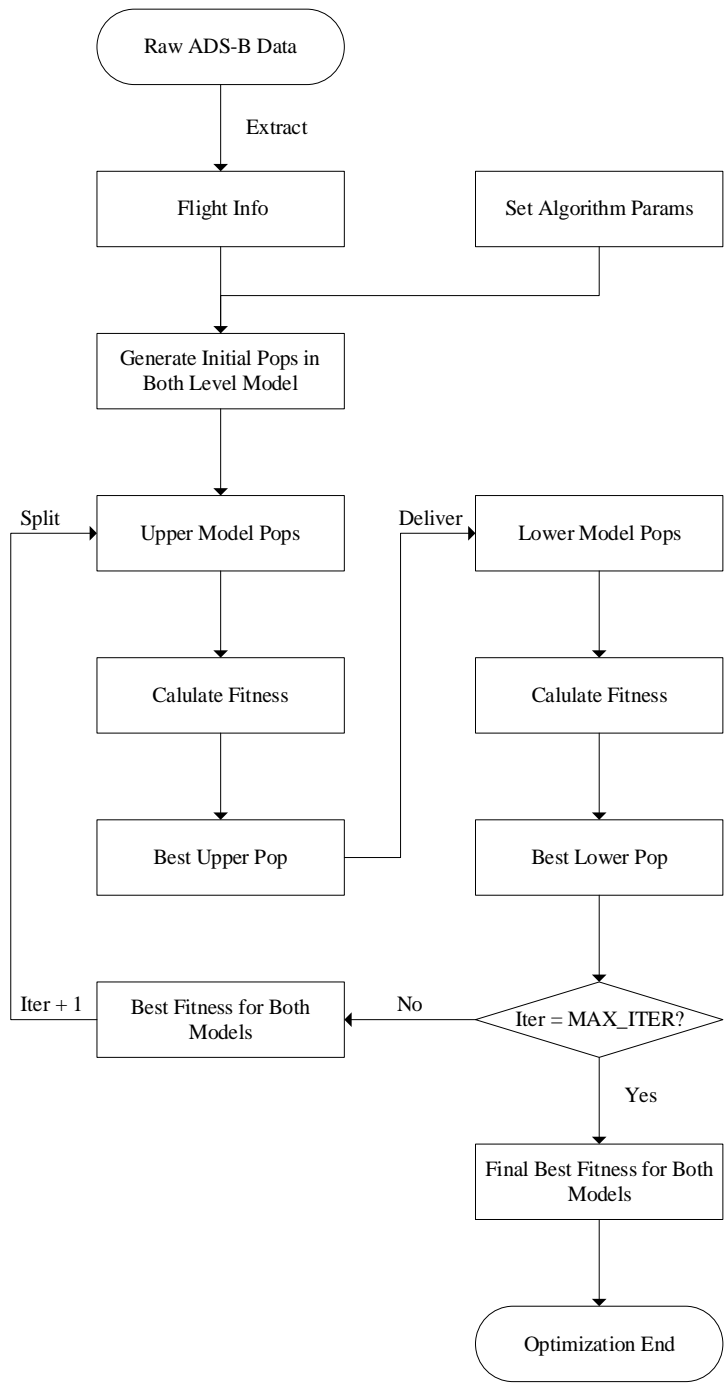

**Fig. 6 Algorithmic Flowchart**

**Step 2** Match the MAS traffic scenarios and set the corresponding optimization objectives.

**Step 3** Set the specific parameters of the algorithm, including the number of upper and lower populations, the number of upper and lower evolutions, the number of co-evolution, mutation, crossover probability, gradient, etc.

**Step 4** According to ETOT, ELDT, and various interval constraints, including the proposed runway time constraints and arrival time constraints, generate the initial upper-level arrival flight population that meets the constraints, representing the arrival time of the arrival flight; the arrival time plus the approach flight time to get the list of landing times. Moreover, the algorithm generates the initial lower-level departure flight population that satisfies the constraints with the upper model and the constraints of the departure flights.

**Step 5** Calculate the fitness value (objective function value) of each individual in the population according to the upper objective function, take out the individual with the highest fitness value, and get the landing time of all flights contained in the individual through the approach flight time simulation, and pass it to the lower model.

**Step 6** The lower model calculates the fitness of all individuals in the population, finds the individual with the highest fitness, and passes it to the upper model;

**Step 7** Update the optimal objective function values of the upper and lower models in the current co-evolution iteration;

**Step 8** If the algorithm reaches the maximum number of iterations, terminate and output the corresponding results; if not, return to Step 5;

## V. Case Study

### A. Case Description

Shanghai TMA was used for the experiments in this paper. There are two airports in the TMA, Shanghai Hongqiao International Airport (ZSSS) and Shanghai Pudong International Airport (ZSPD), where ZSSS is a close-parallel two-runway airport with end-around taxiways, and ZSPD is a four-runway airport consisting of two pairs of close-parallel runways. According to publicly available AIP information, the runways of the two airports are shown in Fig. 7. According to the actual operation, runway 16L/34R and runway 17R/35L at ZSPD airport are used for landing, and runway 16R/34L and runway 17L/35R are used for take-off. Meanwhile, this paper assumes that runway 18L/36R at ZSSS airport is used for landing and runway 18R/36L for take-off to reduce the modeling complexity. The center points of the two airports are 43.98km apart. In summary, the runway configurations[36] used for this case study of the two airports in this paper are shown in Table 1.

Fig. 8 shows the general structure of the airspace of the Shanghai Metroplex terminal area, the location of the airports, the arrival fixes, and the departure fixes. There are five arrival fixes and 10 departure fixes in the TMA. The green and red colors in Fig. 9 show the approach and departure flight trajectories in the TMA after cleaning the

ADS-B data of the TMA for one whole day. The red and blue colors are the boundaries of each sector in the TMA. In actual operation, arrival flights are completely separated from departure flights in the TMA, which is suitable for the experiments of the algorithm proposed in this study.

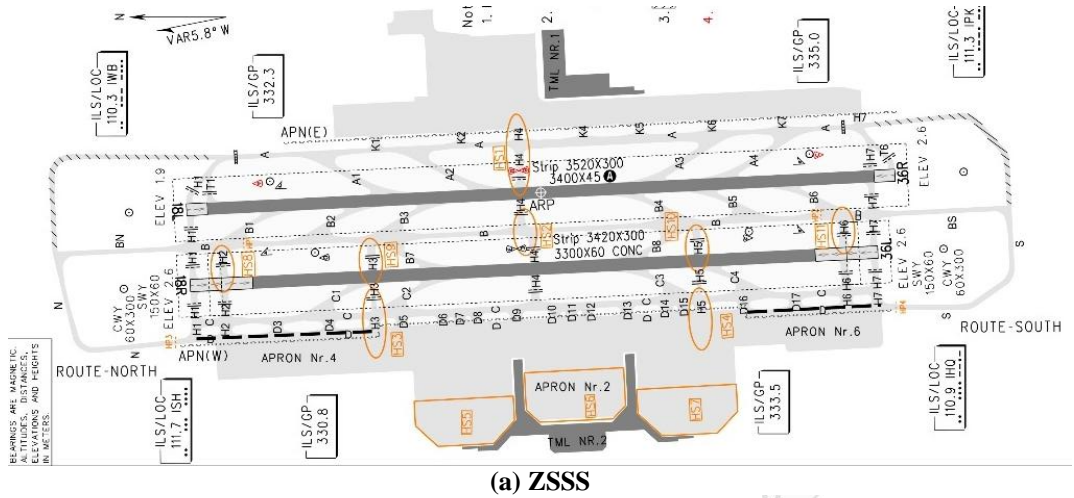

(a) ZSSS

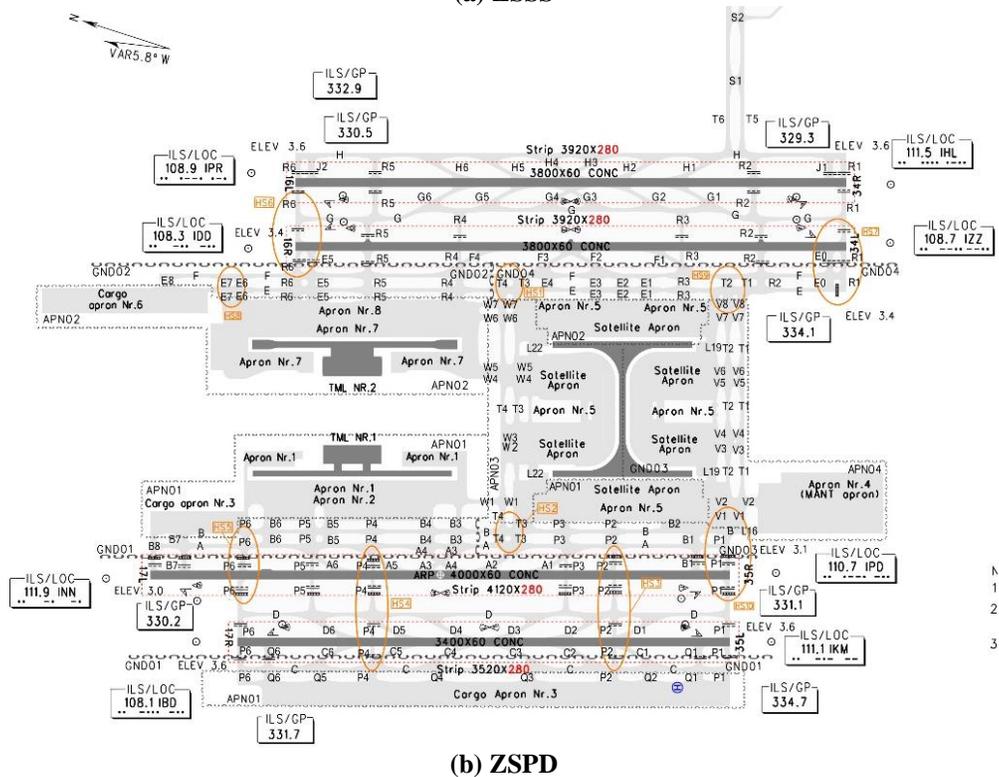

(b) ZSPD

**Fig. 7 Schematic Diagrams of Runways at Two Airports in Shanghai TMA, China**

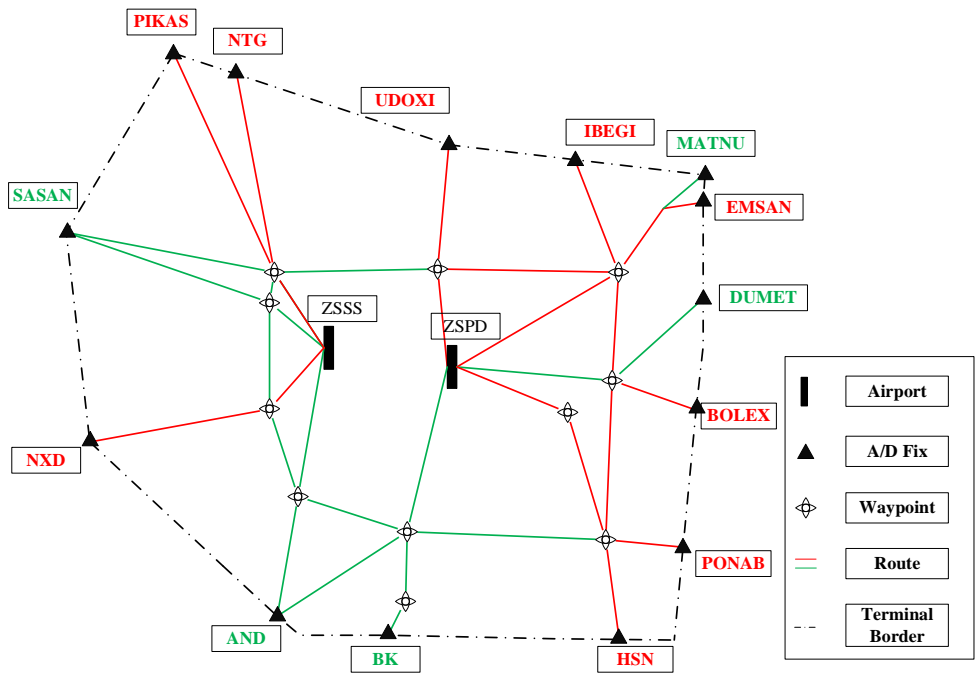

**Fig. 8 Shanghai Metroplex terminal area**

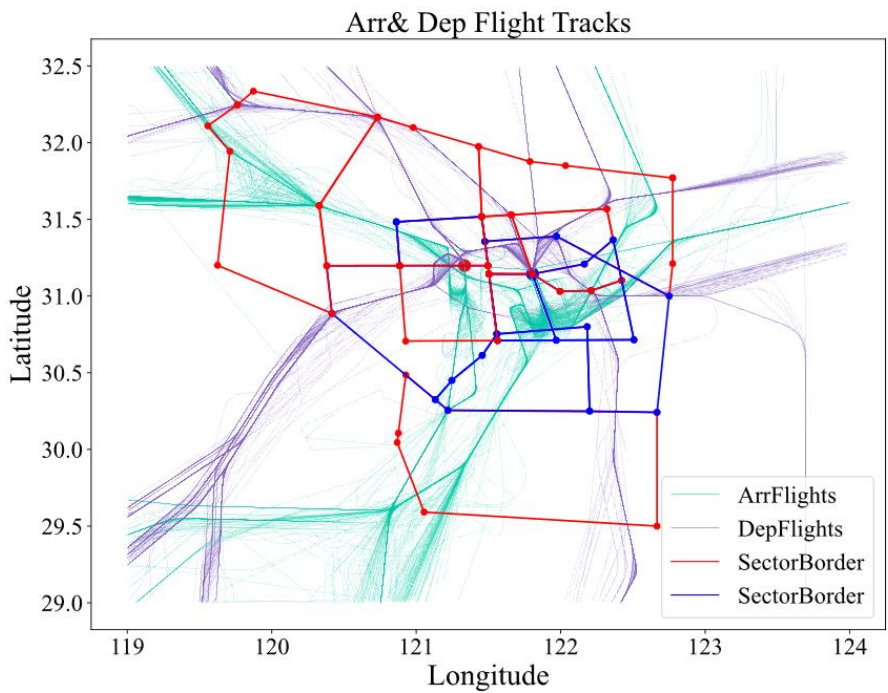

**Fig. 9 Schematic of Typical Daily Trajectories in Shanghai TMA, China**

**B. Experiment Design and Parameters Setting**

*1. Parameters Setting*

According to the ATC separation requirement [28], between different aircraft types, the wake separation of arrival-arrival flights and departure-departure flights are different, as shown in Table 2 and Table 3. Please note that row labels indicate the following aircraft type, and column headers indicate the preceding aircraft type. According to the specific settings of Shanghai TMA, the ATC handover separation for arrival flights is 20km, and that for departure flights is 30km, taking the average flight speed of 800km/h at the handover fixes and converting it into time, the approximate arrival handover separation is 90s, and the departure handover separation is 135s. When an arrival flight precedes a departure flight at the runway, the departure flight shall wait for the arrival flight to vacate from the runway, which takes 45s, and cross the departure runway, which also takes 45s; when a departure flight precedes an arrival flight at the runway, the arrival flight shall wait for the departure flight until the departure flight clear from the runway which spends 45s. Moreover, the maximum position shift is set to 2. Last but not least, the presence of an end-around taxiway at ZSSS means that the crossing runway restriction for a following departure flight only applies when the wingspan of the preceding arrival flight is no less than 36 meters.

**Table 1 Shanghai TMA Runway Configurations**

| Airport | Runway Number | Runway Configuration | Close Parallel Runway |
|---|---|---|---|
| ZSPD | 16L/34R | Arrival | 16R/34L |
| | 16R/34L | Departure | 16L/34R |
| | 17L/35R | Departure | 17R/35L |
| | 17R/35L | Arrival | 17L/35R |
| ZSSS | 18L/36R | Arrival | 18R/36L |
| | 18R/36L | Departure | 18L/36R |

**Table 2 Wake Separation for Arrival Flights**

| Following\Preceding | A380-800 | Heavy | Medium | Light |
|---|---|---|---|---|
| A380-800 | 1min | 2min | 3min | 4min |
| Heavy | 1min | 1min | 2min | 3min |
| Medium | 1min | 1min | 1min | 3min |
| Light | 1min | 1min | 1min | 1min |

**Table 3 Wake Separation for Departure Flights**

| Following\Preceding | A380-800 | Heavy | Medium | Light |
|---|---|---|---|---|
| A380-800 | 1min | 2min | 3min | 3min |
| Heavy | 1min | 1min | 2min | 2min |
| Medium | 1min | 1min | 1min | 2min |
| Light | 1min | 1min | 1min | 1min |

*2. Discussion Based on the MAS Traffic Scenario*

The MAS peak thresholds can be based on the results of [37], where it is determined that MAS enters the peak condition when the traffic reaches 80% of the capacity. Table 4 shows the hourly capacity of the two MAS airports under instrument meteorological conditions in the official document. Based on the previously mentioned peak conditions and Table 4, the peak thresholds of MAS can be calculated as follows:

$(25 \times 2 + 46 \times 2) \times 0.8 = 113.6 (flights/hour)$, similarly, the peak threshold of ZSSS is 40 aircraft/hour; the peak threshold of ZSPD is 73.6 aircraft/hour. Selecting 10 minutes as the unit time for the algorithm to run, the 10-minute peak thresholds for this MAS, ZSSS, and ZSPD are 19, 7, and 12, respectively. Based on these thresholds, the multi-airport system traffic scenarios are experimented for the following situations.

**Table 4 Hourly Capacity of Airports**

| Airport | Hourly Departure Capacity | Hourly Arrival Capacity |
|---|---|---|
| ZSSS | 25 | 25 |
| ZSPD | 46 | 46 |

**Table 5 MAS Traffic Scenario**

| Scenario | MAS | ZSSS | ZSPD |
|---|---|---|---|
| 1 | Peak | Peak | Non-peak |
| 2 | Peak | Non-peak | Peak |
| 3 | Peak | Peak | Peak |
| 4 | Non-peak | Peak | Non-peak |
| 5 | Non-peak | Non-peak | Peak |
| 6 | Non-peak | Non-peak | Non-peak |

Scenario 1) MAS is in peak condition, ZSSS is in peak condition, and ZSPD is not in peak condition.

Scenario 2) MAS is in peak condition, ZSSS is not in peak condition, and ZSPD is in peak condition.

Scenario 3) MAS is in a peak state, ZSSS is in a peak state, and ZSPD is in a peak state.

Scenario 4) MAS is not in a peak state, ZSSS is in a peak state, and ZSPD is not in a peak state.

Scenario 5) MAS is not in a peak state, ZSSS is not in a peak state, and ZSPD is in a peak state.

Scenario 6) MAS is not in peak condition, ZSSS is not in peak condition, and ZSPD is not in peak condition.

The above six traffic scenarios are summarized in Table 5.

## C. Algorithms Performance Comparison

The bi-level algorithms are set as bi-GA, bi-EGA, and bi-SEGA, respectively, and the performance comparison is done under six traffic scenarios to find out the optimal bi-level evolutionary algorithm under various scenarios by testing the convergence of the three algorithms under six different traffic scenarios, respectively. The red dotted line shows the objective function value of the FCFS scheduling strategy in the current traffic scenario, and a smaller objective function represents better performance. This paper compares the FCFS scheduling strategy with the proposed model. Meanwhile, NSGA II was chosen to compare the bi-level genetic algorithm bi-EGA with bi-SEGA. The performance comparison graphs of each algorithm are summarized as shown in Fig. 11 and Fig. 12. In addition, the comparison of the delays of the arrival and departure flights for all algorithms for each scenario is highlighted as shown in Fig. 10 (the underperforming bi-GA is removed to highlight the delay comparison for the remaining scheduling strategies). In the peak scenario Scenario1-3, since the optimization objective is not delay but mainly minimizing the runway occupation time of the departure flights, the bi-level algorithms underperform in terms of delay compared to the FCFS scheduling strategy and NSGA II; in the off-peak hours, the bi-level algorithms start to optimize for delay, and they begin to outperform the FCFS scheduling strategy and NSGA II algorithms in terms of delay. The analysis of the experimental results will be divided into two parts: horizontal comparison and vertical comparison.

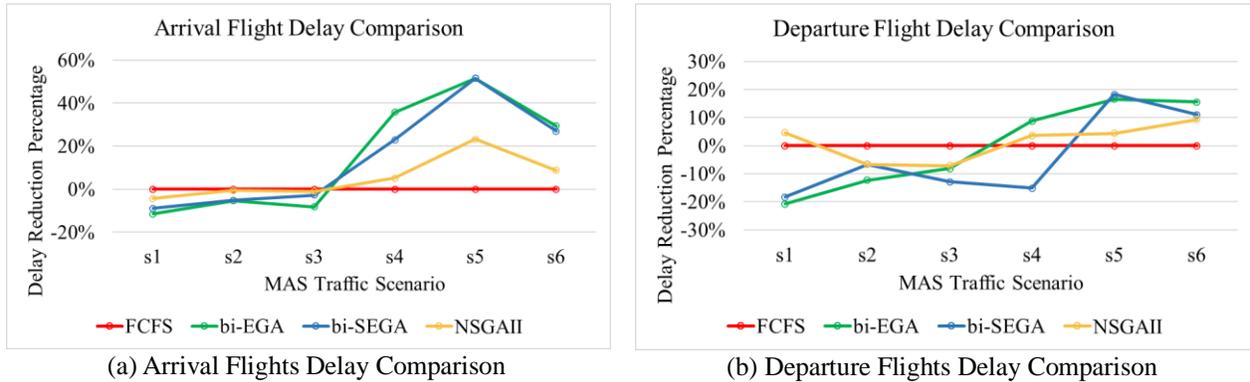

(a) Arrival Flights Delay Comparison    (b) Departure Flights Delay Comparison

**Fig. 10 Algorithms Flight Delay Comparison**

*1. Horizontal Comparison: based on Airport Traffic Peak State*

As shown in Fig. 11(a)-(f), the overall metroplex terminal area is in peak state in Scenario 1-3. In these scenarios, the optimization objective of the arrival flights is to minimize the sequence change, while the optimization objective of the departure flights is to reduce the runway occupation time. In Scenario 1-3, both bi-

EGA and bi-SEGA perform well, with bi-SEGA achieving a significant improvement of 23.7% in reducing the runway occupation time of the departing flights in Scenario 3, which has the highest traffic.

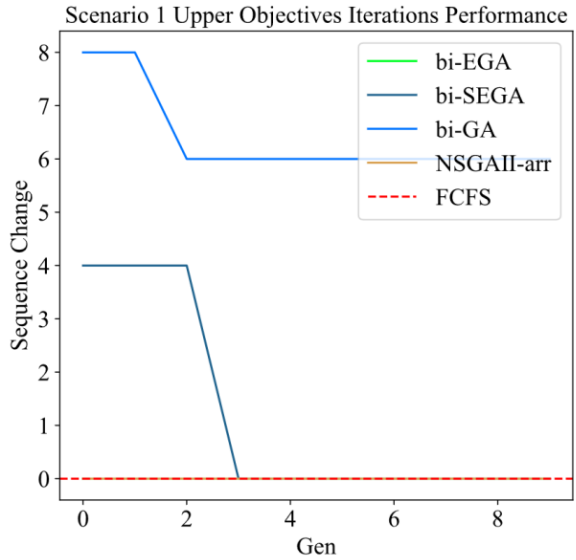

(a) Peak Scenario 1: Upper Level

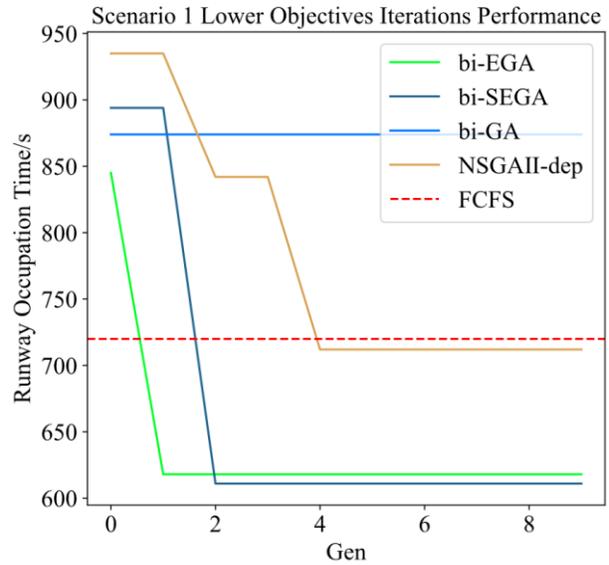

(b) Peak Scenario 1: Lower Level

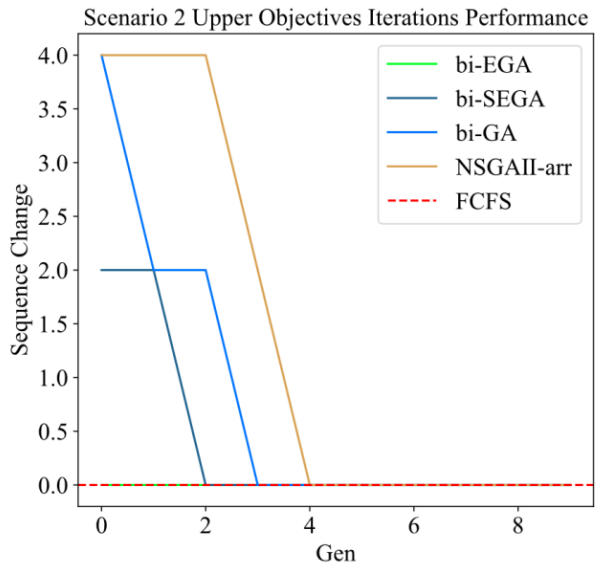

(c) Peak Scenario 2: Upper Level

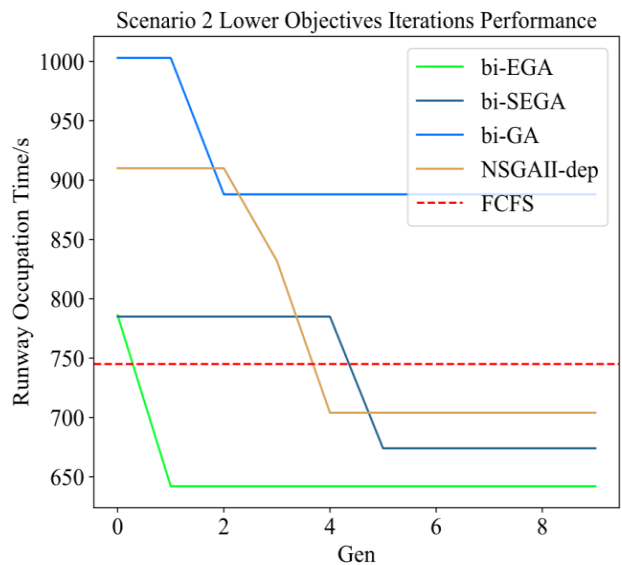

(d) Peak Scenario 2: Lower Level

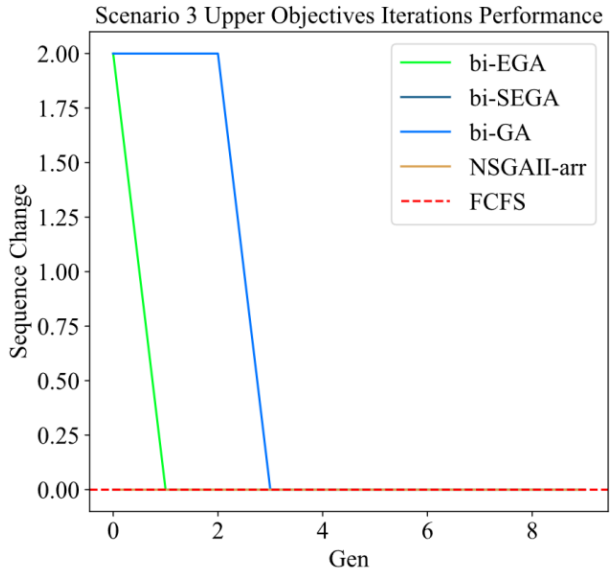 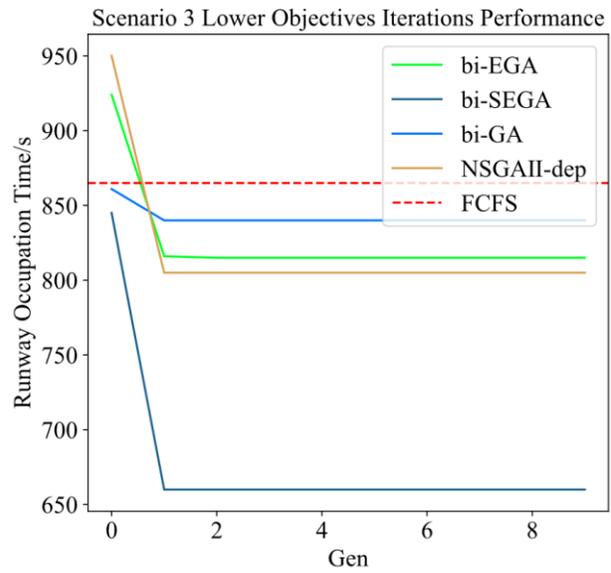

**(e) Peak Scenario 3: Upper Level**  **(f) Peak Scenario 3: Lower Level**

**Fig. 11 Algorithms Performance Comparison at Peak Scenario 1-3**

In Scenario 4-6, the optimization objective for both arrival and departure flights becomes minimizing delays. As shown in Fig. 12(a)-(f), bi-EGA and bi-SEGA still show superior performance, where bi-SEGA reduces the delays of incoming flights by 51.52% while reducing the delays of outgoing flights by 18.05% in Scenario 5. It can be seen that bi-EGA and bi-SEGA can effectively optimize for the defined optimization objective in different traffic scenarios, which reflects its generalization and adaptability.

*2. Vertical comparison: based on algorithm optimization performance and convergence stability*

Regarding algorithm performance, in all traffic scenarios, bi-EGA and bi-SEGA show significantly better optimization performance and convergence stability than the bi-level simple genetic algorithm bi-GA. At the same time, bi-GA performs poorly and fails to outperform the FCFS scheduling strategy in most scenarios, which shows its limitation in dealing with complex sequencing problems. In addition, bi-EGA and bi-SEGA exceed the non-bi-level algorithm NSGAII in terms of optimization performance. Still, in terms of convergence stability, all the experimental scenarios show an overall trend of bi-EGA > NSGA II > bi-SEGA, which reveals that although the bi-SEGA algorithm has superior optimization performance, its convergence stability is not as good as bi-EGA, such as bi-EGA, as well as the non-bi-level multi-objective genetic algorithm NSGA II as a baseline.

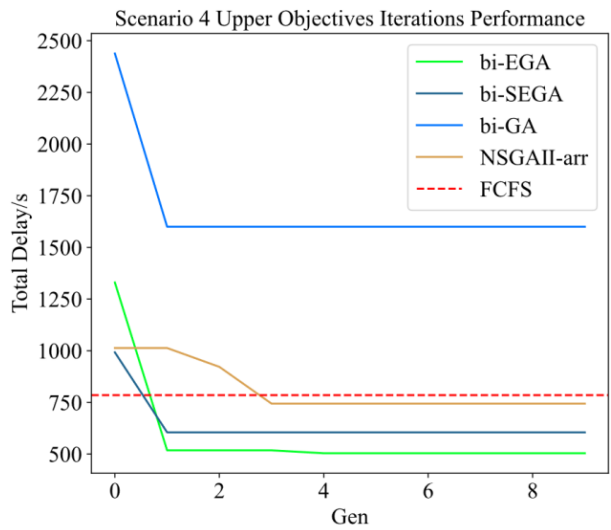

**(a) Non-peak Scenario 4: Upper Level**

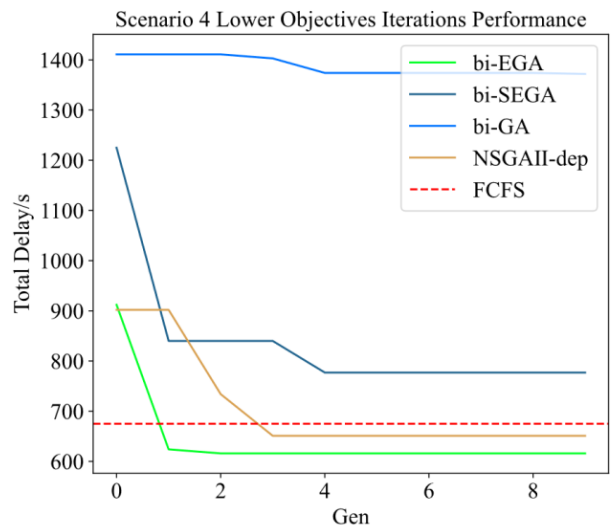

**(a) Non-peak Scenario 4: Upper Level**

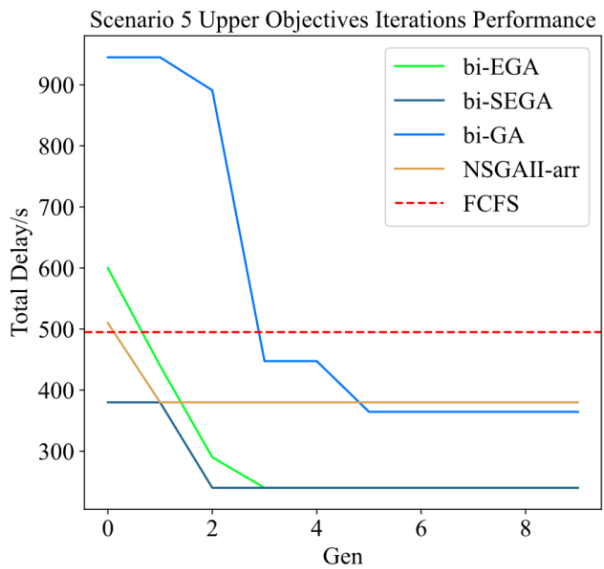

**(c) Non-peak Scenario 5: Upper Level**

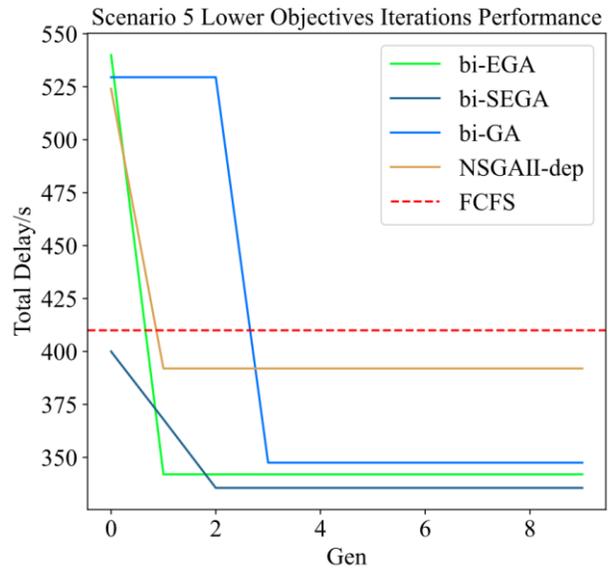

**(d) Non-peak Scenario 5: Lower Level**

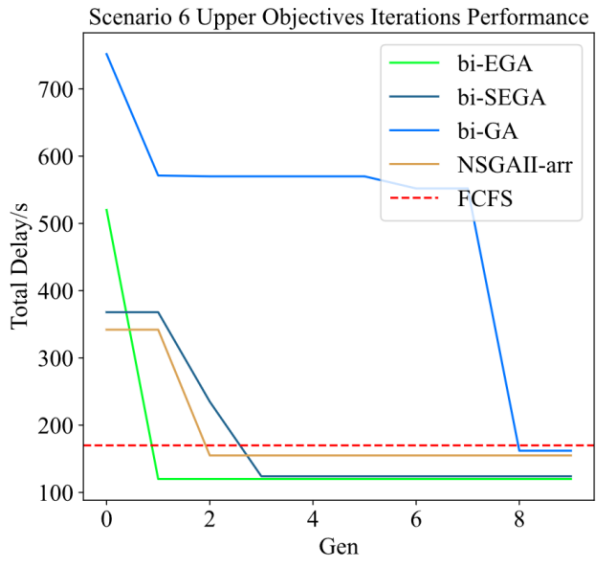 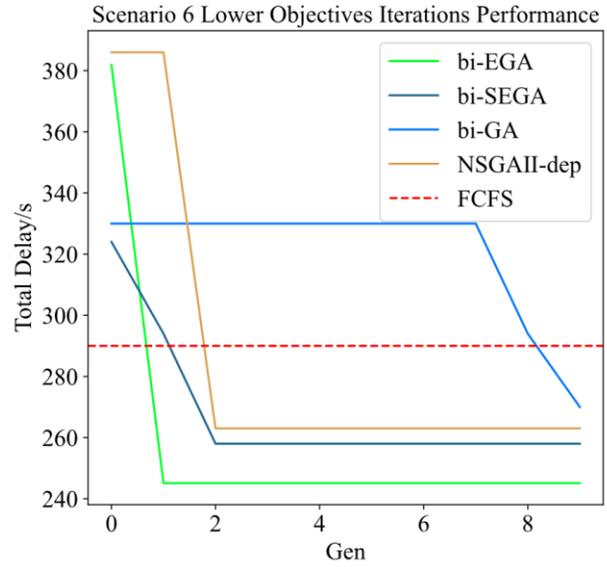

**(e) Non-peak Scenario 6: Upper Level**  **(f) Non-peak Scenario 6: Lower Level**

**Fig. 12 Algorithms Performance Comparison at Non-peak Scenario 4-6**

*3. Summary of Algorithms Performance Comparison*

After the horizontal and vertical comparisons, it can be seen that bi-EGA and bi-SEGA have better convergence performance for various traffic scenarios and can be optimized better than the FCFS strategy and NSGA II algorithm. At the same time, bi-GA performs poorly in most of the scenarios due to its algorithmic limitations. Bi-EGA and bi-SEGA are roughly comparable in their objective functions, each with advantages and disadvantages, and they exhibit significant differences compared to other baseline models, making them well-suited for the scheduling model proposed in this paper.

**D. Ablation Study of Proposed Handover-Fix-Related Improvement**

This paper proposes two improvements related to handover fixes: the Constant Relative Sequence for Same-Path Flights constraint, referred to in this section as CRSSPF, and the Staggered Assignment of Handover Altitudes, referred to as SAHA. To analyze the improvement brought by the two proposed handover-fix-related improvements, we selected the algorithm bi-EGA for ablation analysis for the proposed model, the model without Improvement 1, the model without Improvement 2, and the model without both improvements, as shown in Fig. 13. The ablation study is divided into two parts: one focusing on MAS peak scenarios, as depicted in Fig. 13(a) and Fig. 13(b), and the other on MAS non-peak scenarios, as illustrated in Fig. 13(c) and Fig. 13(d).

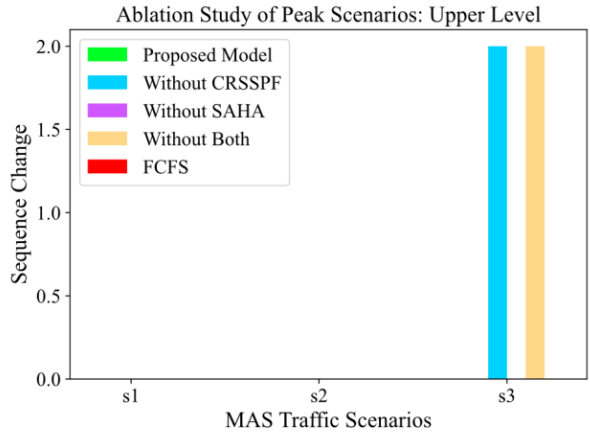
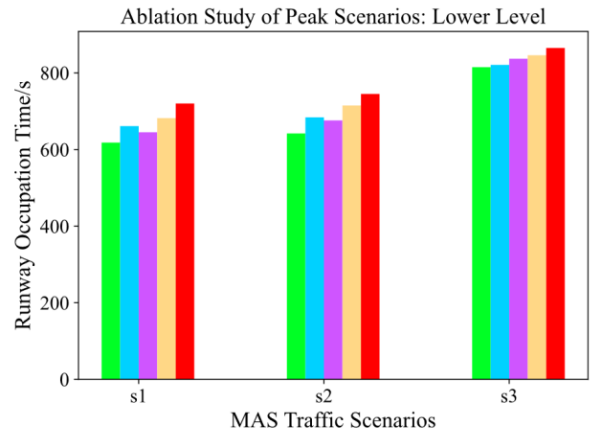

**(a) Peak Scenarios 1-3: Upper Level**     **(b) Peak Scenarios 1-3: Lower Level**

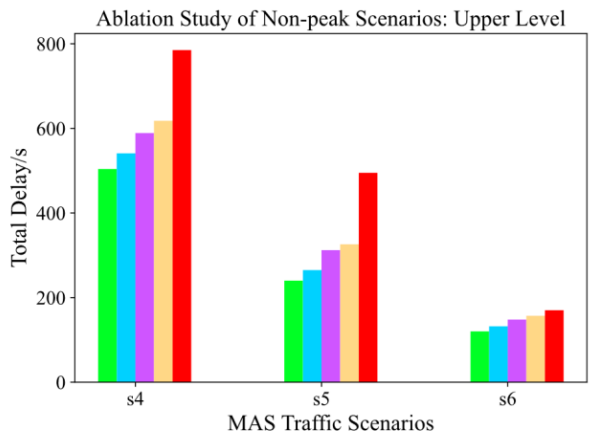
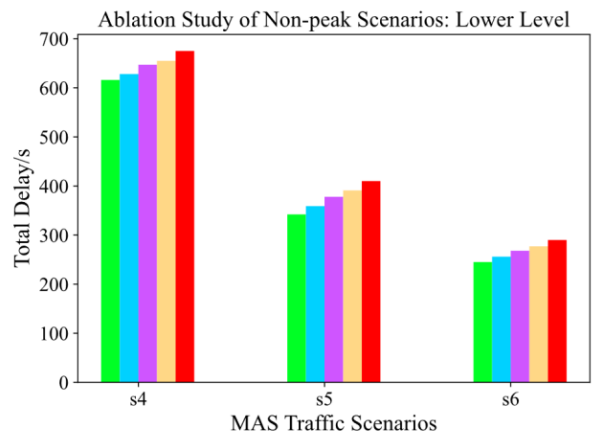

**(c) Non-peak Scenarios 4-6: Upper Level**     **(d) Non-peak Scenarios 4-6: Lower Level**

**Fig. 13 Ablation Study of Proposed Handover-Fix-Related Improvements**

1. *Ablation Study of Peak Scenarios 1-3*

As illustrated in Fig. 13(a) and Fig. 13(b), the proposed model exhibited the best performance across all three scenarios within the peak conditions of the Metroplex Terminal Area. In Scenario 1 and Scenario 2, the upper-level models optimizing arrival flights successfully maintained the sequence of arrival flights across all configurations. However, in the lower-level models focusing on departure flights, models lacking the Constant Relative Sequence for Same-Path Flights (CRSSPF) constraint performed worse than those lacking the Staggered Assignment of Handover Altitudes (SAHA). This finding underscores the critical importance of CRSSPF in optimizing runway occupation time for departure flights.

In Scenario 3, which represents the highest traffic volume scenario, the lower-level models lacking CRSSPF outperformed those lacking SAHA. This indicates that SAHA also significantly optimizes runway occupation time for departure flights under higher traffic conditions. Furthermore, in the upper-level models of Scenario 3, both the models lacking CRSSPF and those lacking both improvements exhibited sequence changes in arrival flights. This demonstrates the essential role of CRSSPF in maintaining the stability of arrival flight sequences in high-traffic scenarios.

*2. Ablation Study of Non-peak Scenarios 4-6*

As depicted in Fig. 13(c) and Fig. 13(d), within the Metroplex Terminal Area's non-peak scenarios, the upper-level model shifts its objective to minimizing arrival flight delays through sequencing optimization. In contrast, the lower-level model focuses on optimizing departure flight delays. The Proposed Model consistently outperformed the other models, thereby validating the effectiveness of both the CRSSPF and the SAHA in reducing flight delays. Furthermore, in the comparative analysis between the two improvements, models lacking SAHA demonstrated inferior performance compared to those lacking CRSSPF. This finding highlights the more significant role of SAHA in minimizing flight delays relative to CRSSPF, underscoring the critical importance of staggered assignment in enhancing overall scheduling efficiency.

*3. Summary of Ablation Study*

The ablation study assesses the contributions of the CRSSPF and SAHA within peak and non-peak scenarios in the Metroplex Terminal Area. Results demonstrate that the proposed model, which integrates CRSSPF and SAHA, consistently outperforms models that lack improvements across all scenarios. Specifically, CRSSPF is essential for maintaining the stability of arrival flight sequences and optimizing runway occupation time during peak traffic conditions. At the same time, SAHA plays a more significant role in minimizing flight delays in non-peak scenarios. These findings highlight the complementary effectiveness of CRSSPF and SAHA in enhancing overall scheduling efficiency.

## VI. Conclusion

This paper proposes a collaborative sequencing model based on bi-level planning for arrival and departure flights in metroplex terminal areas. We use bi-level heuristic evolutionary algorithms to solve the problem. The bi-level planning design ensures that the proposed model gives priority to the optimization of approaching flights; in different traffic scenarios of MAS, corresponding optimization objectives are set to optimize the approaching flights and departing flights, respectively, to achieve the optimization of dynamic sequencing in different periods of metroplex terminal areas. When the MAS is in peak hours, the bi-level planning model pays more attention to reducing the controller load; when the MAS is in off-peak hours, the bi-level planning model pays more attention to reducing flight delays. In addition, in this paper, the role of handover fixes as entrances and exits in the TMA is profoundly discussed in the model design. Two constraints, namely, flexible handover altitude and constant relative order of same-path flights, are proposed. The bi-level heuristic evolutionary algorithms bi-EGA and bi-SEGA outperform the FCFS scheduling strategy and the non-bi-level NSGA II algorithm in performance; bi-EGA and bi-SEGA are suitable algorithms for solving the proposed problem.

This paper addresses the research gap regarding the in-depth exploration of handover fixes in flight sequencing for arrival and departure flights within metroplex terminal areas. Additionally, it provides a deep analysis of the varying demands of arrival and departure flights under different traffic scenarios. The result of this paper can serve as a booster for the actual operation of air traffic control in metroplex terminal areas. Future research can develop a bi-level multi-objective heuristic evolutionary algorithm suitable for the flight sequencing problem of arrival and departure flights in metroplex terminal areas to cope with the complex demands of all parties in CDM.

## Funding Sources

This work was supported by the National Key R&D Program of China (No. 2022YFB4300905) and the Graduate Research and Innovation Projects of Jiangsu Province (No. KYCX24 0598).

## References

[1] Federal Aviation Administration, "2018-2019 NextGen Implementation Plan," Washington DC,USA, 2019.

[2] ICAO, "Working Document for the Aviation System Block Upgrades–The Framework for Global Harmonization," Montreal,Canada, 2013.

[3] SESAR, "European ATM Master Plan – The Roadmap for Delivering High Performing Aviation for Europe, Edition 2015," Brussels, Belgium, 2015.

[4]     Wang, L., Ding, Q., and Wei, D., "Based on Point Merge for Paired Approach Sequencing on Closely Spaced Parallel Runways," *Journal of Traffic and Transportation Engineering (English Edition)*, Vol. 10, No. 5, 2023, pp. 934–946. https://doi.org/10.1016/j.jtte.2023.09.001

[5]     Ma Y., Hu M., Zhang H., Yin J., and Wu F., "Optimized method for collaborative arrival sequencing and scheduling in metroplex terminal area," *Acta Aeronautica et Astronautica Sinica*, Vol. 36, No. 07, 2015, pp. 2279–2290. https://doi.org/10.7527/s1000-6893.2014.0280

[6]     Jiang, H., Liu, J., and Zhou, W., "Bi-level Programming Model for Joint Scheduling of Arrival and Departure Flights Based on Traffic Scenario," *Transactions of Nanjing University of Aeronautics and Astronautics*, Vol. 38, No. 4, 2021, p. 671-684. http://dx.doi.org/10.16356/j.1005-1120.2021.04.013

[7]     Jiang, H., Zeng, W., Wei, W., and Tan, X., "A Bilevel Flight Collaborative Scheduling Model with Traffic Scenario Adaptation: An Arrival Prior Perspective," *Computers & Operations Research*, Vol. 161, 2024, p. 106431. https://doi.org/10.1016/j.cor.2023.106431

[8]     DEAR, R. G., "THE DYNAMIC SCHEDULING OF AIRCRAFT IN THE NEAR TERMINAL AREA.," 1976.

[9]     Hu, X.-B., and Chen, W.-H., "Receding Horizon Control for Aircraft Arrival Sequencing and Scheduling," *IEEE Transactions on Intelligent Transportation Systems*, Vol. 6, No. 2, 2005, pp. 189–197. https://doi.org/10.1109/TITS.2005.848365

[10]    Xiao-Bing Hu, and Di Paolo, E., "Binary-Representation-Based Genetic Algorithm for Aircraft Arrival Sequencing and Scheduling," *IEEE Transactions on Intelligent Transportation Systems*, Vol. 9, No. 2, 2008, pp. 301–310. https://doi.org/10.1109/TITS.2008.922884

[11]    Hu, X., and Di Paolo, E., "An Efficient Genetic Algorithm with Uniform Crossover for Air Traffic Control," *Computers & Operations Research*, Vol. 36, No. 1, 2009, pp. 245–259. https://doi.org/10.1016/j.cor.2007.09.005

[12]    Salehipour, A., Modarres, M., and Moslemi Naeni, L., "An Efficient Hybrid Meta-Heuristic for Aircraft Landing Problem," *Computers & Operations Research*, Vol. 40, No. 1, 2013, pp. 207–213. https://doi.org/10.1016/j.cor.2012.06.004

[13]   Cao, Y., Rathinam, S., and Sun, D., "Greedy-Heuristic-Aided Mixed-Integer Linear Programming Approach for Arrival Scheduling," *Journal of Aerospace Information Systems*, Vol. 10, No. 7, 2013, pp. 323–336. https://doi.org/10.2514/1.I010030

[14]   Zhang, J., Zhao, P., Zhang, Y., Dai, X., and Sui, D., "Criteria Selection and Multi-Objective Optimization of Aircraft Landing Problem," *Journal of Air Transport Management*, Vol. 82, 2020, p. 101734. https://doi.org/10.1016/j.jairtraman.2019.101734

[15]   Jiang, F., and Zhang, Z., "Optimal Sequencing of Arrival Flights at Metroplex Airports: A Study on Shared Waypoints Based on Path Selection and Rolling Horizon Control," *Aerospace*, Vol. 10, No. 10, 2023, p. 881. https://doi.org/10.3390/aerospace10100881

[16]   Murça, M., and Müller, C., "Control-Based Optimization Approach for Aircraft Scheduling in a Terminal Area with Alternative Arrival Routes," *Transportation Research Part E-Logistics and Transportation Review*, Vol. 73, 2015, pp. 96–113. https://doi.org/10.1016/j.tre.2014.11.004

[17]   Wang, Y., "Departure Scheduling in a Multi-Airport System," presented at the Eighth USA/Europe Air Traffic Management Research and Development Seminar, 2009.

[18]   Montoya, J., Rathinam, S., and Wood, Z., "Multiobjective Departure Runway Scheduling Using Dynamic Programming," *IEEE Transactions on Intelligent Transportation Systems*, Vol. 15, No. 1, 2014, pp. 399–413. https://doi.org/10.1109/TITS.2013.2283256

[19]   Ming Liu, Zhihui Sun, Xiaoning Zhang, and Feng Chu, "A Two-Stage No-Wait Hybrid Flow-Shop Model for the Flight Departure Scheduling in a Multi-Airport System," presented at the 2017 IEEE 14th International Conference on Networking, Sensing and Control (ICNSC), Calabria, Italy, 2017. https://doi.org/10.1109/ICNSC.2017.8000142

[20]   Zhong, H., Guan, W., Zhang, W., Jiang, S., and Fan, L., "A Bi-Objective Integer Programming Model for Partly-Restricted Flight Departure Scheduling," *PLOS One*, Vol. 13, No. 5, 2018, p. e0196146. https://doi.org/10.1371/journal.pone.0196146

[21]   Sandamali, G. G. N., Su, R., and Zhang, Y., "Flight Routing and Scheduling Under Departure and En Route Speed Uncertainty," *IEEE Transactions on Intelligent Transportation Systems*, Vol. 21, No. 5, 2020, pp. 1915–1928. https://doi.org/10.1109/TITS.2019.2907058

[22]   Ma, J., Sbihi, M., and Delahaye, D., "Optimization of Departure Runway Scheduling Incorporating Arrival Crossings," *International Transactions in Operational Research*, Vol. 28, No. 2, 2021, pp. 615–637. https://doi.org/10.1111/itor.12657

[23]   Chandrasekar, S., and Hwang, I., "Algorithm for Optimal Arrival and Departure Sequencing and Runway Assignment," *Journal of Guidance, Control, and Dynamics*, Vol. 38, No. 4, 2015, pp. 601–613. https://doi.org/10.2514/1.G000084

[24]   Sölveling, G., and Clarke, J.-P., "Scheduling of Airport Runway Operations Using Stochastic Branch and Bound Methods," *Transportation Research Part C: Emerging Technologies*, Vol. 45, 2014, pp. 119–137. https://doi.org/10.1016/j.trc.2014.02.021

[25]   Miaotian, S., Junfeng, Z., Tengteng, G., and Zhixiang, Z., "A Combined Arrival and Departure Scheduling for Multi-Airport System," *Transactions of Nanjing University of Aeronautics and Astronautics*, Vol. 34, No. 05, 2017, pp. 578–585. https://doi.org/10.16356/j.1005-1120.2017.05.578

[26]   Ahmed, M. S., Alam, S., and Barlow, M., "A Cooperative Co-Evolutionary Optimisation Model for Best-Fit Aircraft Sequence and Feasible Runway Configuration in a Multi-Runway Airport," *Aerospace*, Vol. 5, No. 3, 2018, p. 85. https://doi.org/10.3390/aerospace5030085

[27]   Yang, H., Buire, C., Delahaye, D., and Le, M., "A Heuristic-Based Multi-Objective Flight Schedule Generation Framework for Airline Connectivity Optimisation in Bank Structure: An Empirical Study on Air China in Chengdu," *Journal of Air Transport Management*, Vol. 116, 2024, p. 102571. https://doi.org/10.1016/j.jairtraman.2024.102571

[28]   Yang, H., Delahaye, D., O'Connell, J. F., and Le, M., "Enhancing Airline Connectivity: An Optimisation Approach for Flight Scheduling in Multi-Hub Networks with Bank Structures," *Transportation Research Part E: Logistics and Transportation Review*, Vol. 191, 2024, p. 103715. https://doi.org/10.1016/j.tre.2024.103715

[29]   Abdelghany, A., Abdelghany, K., and Guzhva, V. S., "Schedule-Level Optimization of Flight Block Times for Improved Airline Schedule Planning: A Data-Driven Approach," *Journal of Air Transport Management*, Vol. 115, 2024, p. 102535. https://doi.org/10.1016/j.jairtraman.2023.102535

[30] Wen, X., Sun, X., Ma, H.-L., and Sun, Y., "A Column Generation Approach for Operational Flight Scheduling and Aircraft Maintenance Routing," *Journal of Air Transport Management*, Vol. 105, 2022, p. 102270. https://doi.org/10.1016/j.jairtraman.2022.102270

[31] Deb, K., and Sinha, A., "Constructing Test Problems for Bilevel Evolutionary Multi-Objective Optimization," presented at the 2009 IEEE Congress on Evolutionary Computation (CEC), Trondheim, Norway, 2009. https://doi.org/10.1109/CEC.2009.4983076

[32] "Civil Aviation Air Traffic Management Rules (CCAR-93TM-R6)," Ministry of Transport of the People's Republic of China, Beijing, 2023.

[33] Chen, T., and Guestrin, C., "XGBoost: A Scalable Tree Boosting System," presented at the Proceedings of the 22nd ACM SIGKDD International Conference on Knowledge Discovery and Data Mining, New York, NY, USA, 2016. https://doi.org/10.1145/2939672.2939785

[34] Yang, C., Zhang, J., Gui, X., Peng, Z., and Wang, B., "A Data-Driven Method for Flight Time Estimation Based on Air Traffic Pattern Identification and Prediction," *Journal of Intelligent Transportation Systems*, Vol. 28, No. 3, 2024, pp. 352–371. https://doi.org/10.1080/15472450.2022.2130693

[35] Ye B., Bao X., Liu B., and Tian Y., "Machine learning for aircraft approach time prediction," *Acta Aeronautica et Astronautica Sinica*, Vol. 41, No. 10, 2020, pp. 359–370. https://doi.org/10.7527/S10006893.2020.24136

[36] Wang, S., Zeng, W., Jiang, H., and Tan, X., "Mining Airport Runway Configurations from Flight Trajectories," *Journal of Aerospace Information Systems*, Vol. 21, No. 3, 2024, pp. 290–293. https://doi.org/10.2514/1.I011321

[37] Zhao, X., Wang, Y., Li, L., and Delahaye, D., "A Queuing Network Model of a Multi-Airport System Based on Point-Wise Stationary Approximation," *Aerospace*, Vol. 9, No. 7, 2022. https://doi.org/10.3390/aerospace9070390